\documentclass[letterpaper, 10pt, conference]{ieeeconf}
\IEEEoverridecommandlockouts
\usepackage{cite}
\usepackage{amsmath,amssymb,amsfonts}
\usepackage{graphicx}
\usepackage{textcomp}
\usepackage{xcolor}
\usepackage{booktabs}
\usepackage{multirow}
\usepackage{url}
\usepackage{subcaption}
\usepackage{hyperref}
\usepackage{algorithm}
\usepackage{algorithmic}
\usepackage{float}
\usepackage{stfloats}

\begin{document}

\title{\LARGE \bf Multi-Agent Reinforcement Learning for Safe Autonomous Driving Under Pedestrian Behavioral Uncertainty}

\author{Prakash Aryan$^{1}$, Kaushik Raghupathruni$^{1}$, Timo Kehrer$^{1}$, and Sebastiano Panichella$^{1,2}$%
\thanks{This work is supported by the SNSF SwarmOps project (No.\ 200021\_219732).}%
\thanks{Code: \url{https://github.com/prakash-aryan/marl-sdc-pedestrian-uncertainty}}%
\thanks{$^{1}$University of Bern, Bern, Switzerland. {\tt\small \{prakash.aryan, timo.kehrer, sebastiano.panichella\}@unibe.ch}, {\tt\small kaushik.raghupathruni@students.unibe.ch}}%
\thanks{$^{2}$AI4I, The Italian Institute of Artificial Intelligence, Turin, Italy. {\tt\small sebastiano.panichella@ai4i.it}}%
}

\maketitle
\thispagestyle{empty}
\pagestyle{empty}

\begin{abstract}
Simulation-based testing of self-driving cars (SDCs) typically relies on scripted pedestrian models that do not capture the heterogeneity and uncertainty of real crossing behavior, limiting the realism of safety assessments, especially for jaywalking, which is governed by latent personality traits the vehicle cannot observe. We hypothesize that jointly training pedestrians and the SDC with multi-agent reinforcement learning (MARL) yields more realistic interaction scenarios than training against fixed pedestrian policies, and that the behavior gap between predictable and unpredictable crossings can be measured directly from trajectories. We co-train an SDC and 12 pedestrians using Multi-Agent Proximal Policy Optimization (MAPPO): pedestrian locomotion follows scripted Dijkstra pathfinding while an RL policy controls high-level go/wait decisions, and jaywalking probability depends on a per-pedestrian trait sampled at episode start and hidden from the SDC. In 500-episode evaluations, the co-trained SDC reached 78\% of goals with a 14\% collision rate, versus 35\%/33\% for the best rule-based baseline. A speed differential metric shows the SDC traveled 2.65\,m/s faster near jaywalkers than near crosswalk users at close range (0--3\,m), indicating jaywalking encounters were not anticipated. Jaywalking was 13\% of crossing events but 62\% of collisions, and co-training reduced collisions by 30\% relative to single-agent RL as pedestrians learned to wait when the SDC approached at speed.
\end{abstract}

\section{Introduction}
    Pedestrian-vehicle interactions are a recurring source of urban collisions, and self-driving cars (SDCs) must account for them during training and evaluation. Existing simulation-based testing of SDCs typically relies on scripted pedestrian motion or simplified crossing rules that do not reflect the heterogeneity of real human behavior~\cite{rashid_simulation_2024, birchler_how_2024}, raising a question for SDC assessment: are we training and testing SDCs against pedestrian models realistic enough to expose latent-intent situations such as jaywalking? We hypothesize that co-training pedestrians and the SDC with multi-agent reinforcement learning (MARL), with a personality-driven jaywalking mechanism invisible to the vehicle, yields more realistic interaction scenarios than single-agent training, and that the gap between predictable and unpredictable crossings can be measured from trajectories. We instantiate this idea with Multi-Agent Proximal Policy Optimization (MAPPO)~\cite{yu_surprising_2022}.

Prior work on pedestrian-vehicle interaction has examined risk-aware RL for jaywalker encounters~\cite{zhang_decision-making_2025}, social force based simulation~\cite{helbing_social_1995, rashid_simulation_2024}, and survey evidence on jaywalking near automated vehicles~\cite{dong_will_2024}. MARL has been applied to cooperative driving at intersections~\cite{yu_multi-agent_2025, zhang_multi-agent_2024}. However, these works either fix the pedestrian model or train the SDC in isolation; none co-train pedestrians and the SDC with explicit, trait-driven behavioral uncertainty.

\begin{figure}[!htb]
\centering
\begin{subfigure}[b]{0.48\columnwidth}
\includegraphics[width=\linewidth]{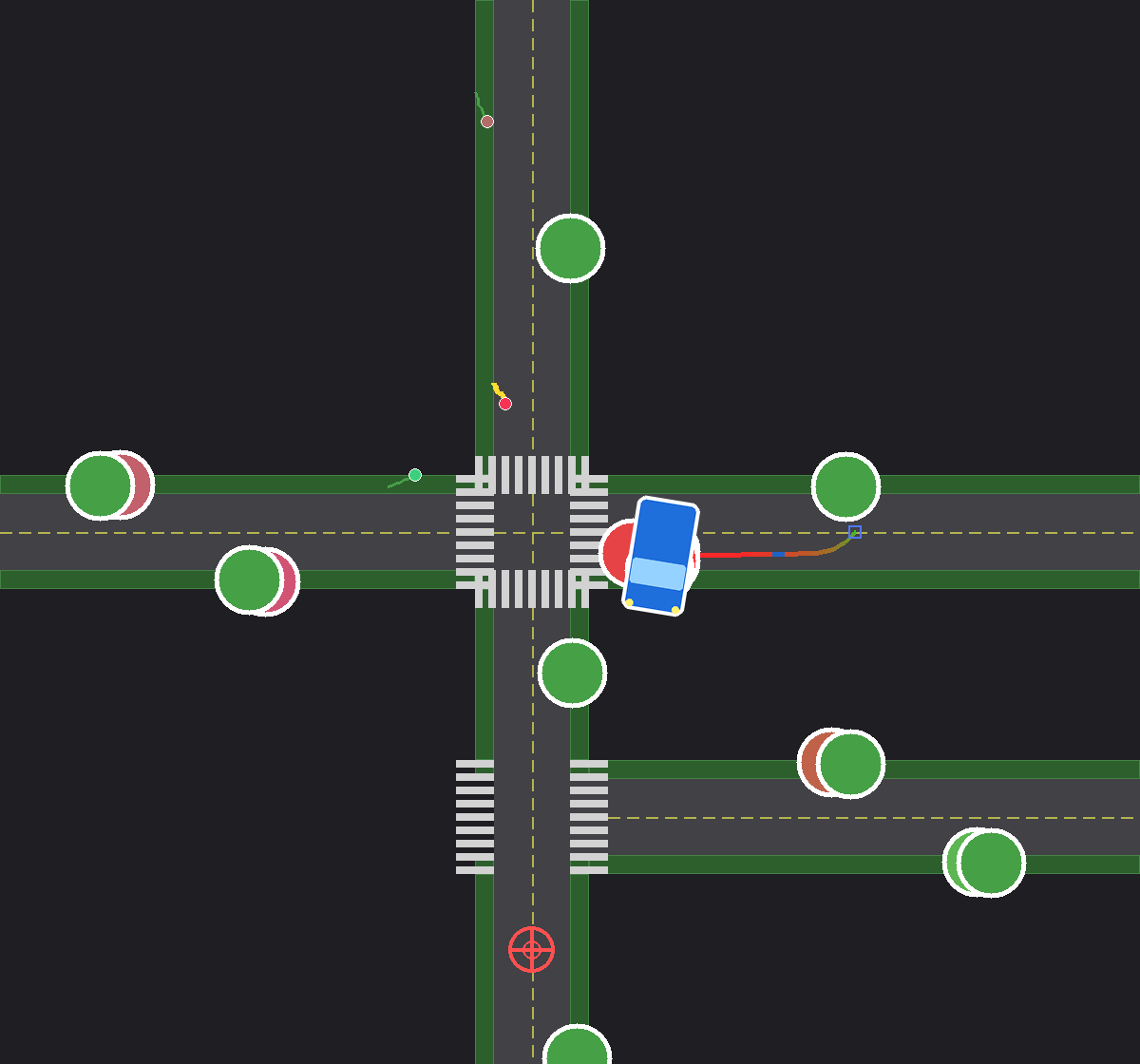}
\caption{Collision with jaywalker}
\label{fig:intro_collision}
\end{subfigure}
\hfill
\begin{subfigure}[b]{0.48\columnwidth}
\includegraphics[width=\linewidth]{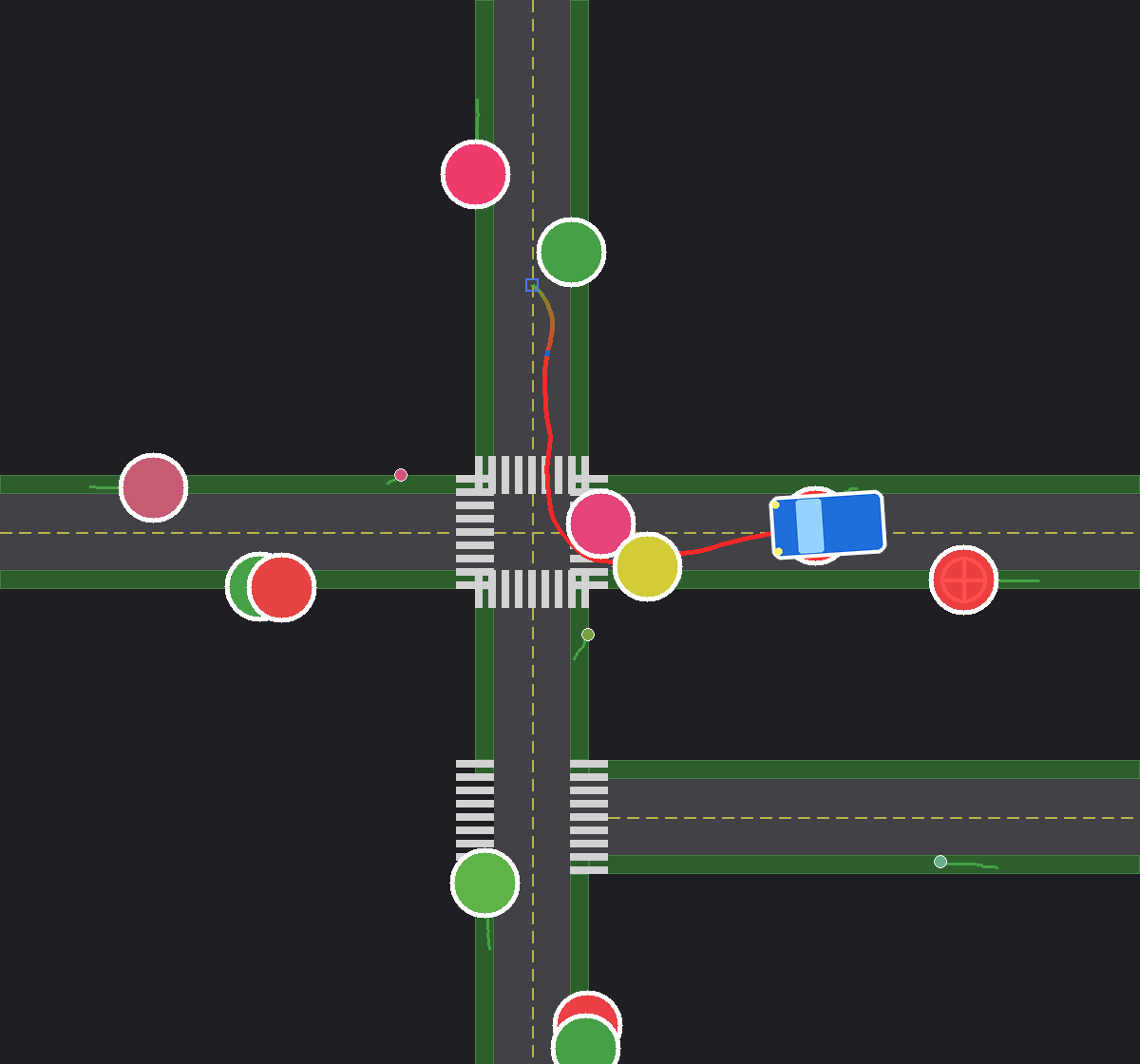}
\caption{Successful avoidance}
\label{fig:intro_avoid}
\end{subfigure}
\caption{Two outcomes in our environment. The blue rectangle is the SDC; coloured dots are pedestrians (hue encodes jaywalking tendency, green cautious to red reckless; yellow segments mark active jaywalking); grey is road, green strips are sidewalks, and the red crosshair is the goal. (a) Collision with a jaywalker. (b) The SDC steers around a jaywalker and reaches its goal. A collision is registered when the SDC and a pedestrian centre come within 1.5\,m.}
\label{fig:intro_scenarios}
\end{figure}

This paper makes three contributions. \emph{First}, a MARL environment for pedestrian-SDC co-training in which jaywalking is driven by a latent personality trait hidden from the SDC, extending prior MARL intersection work that uses at most a handful of scripted pedestrians~\cite{yu_multi-agent_2025}. \emph{Second}, a speed differential metric that quantifies, from trajectories, how the SDC responds differently to predictable crosswalk and unpredictable jaywalking encounters. \emph{Third}, empirical evidence that co-training produces emergent cooperative waiting and reduces collisions by 30\% relative to single-agent RL, even when the single-agent SDC is later paired with the MARL pedestrian policy.

\section{Related Work}

\textbf{Multi-agent RL.} Cooperative MARL methods include MADDPG~\cite{lowe_multi-agent_2017}, QMIX~\cite{rashid_monotonic_2020}, COMA~\cite{foerster_counterfactual_2018}, and IPPO~\cite{witt_is_2020}. MAPPO~\cite{yu_surprising_2022}, built on proximal policy optimization (PPO)~\cite{schulman_proximal_2017} with generalized advantage estimation (GAE)~\cite{schulman_high-dimensional_2018}, uses the centralized training with decentralized execution (CTDE) paradigm~\cite{oliehoek_concise_2016}: a centralized critic during training, decentralized actors at execution. We adopt MAPPO with a shared centralized critic.

\textbf{Pedestrian behavior.} The social force model~\cite{helbing_social_1995} is foundational for pedestrian simulation; personality traits have been added to heterogeneous models~\cite{xue_fuzzy_2017}, and recent work integrates sensory-motor constraints into RL-based pedestrian policies~\cite{wang_modeling_2024}. Trajectory prediction methods such as Social LSTM~\cite{alahi_social_2016} and Trajectron++~\cite{salzmann_trajectron_2021} forecast paths but do not model crossing decisions. Khuzam et al.~\cite{khuzam_impact_2025} studied jaywalking via Markov games, and Zhang et al.~\cite{zhang_decision-making_2025} proposed risk-aware RL for jaywalker interactions with the SDC trained in isolation.

\textbf{Uncertainty and simulation testing.} Kendall and Gal~\cite{kendall_what_2017} distinguish aleatoric and epistemic uncertainty, Hoel et al.~\cite{hoel_tactical_2020} apply uncertainty estimation to tactical driving, and Wang et al.~\cite{wang_uncertainty_2025} survey uncertainty quantification (UQ) methods for autonomous vehicles. Standard SDC testbeds include CARLA~\cite{dosovitskiy_carla_2017} and SUMO~\cite{lopez_microscopic_2018}. Birchler et al.~\cite{birchler_how_2024} show simulation-based SDC testing does not always align with human perception of realism, motivating richer pedestrian models. GPU-accelerated RL via JAX~\cite{bradbury_jax_2018} is demonstrated by JaxMARL~\cite{rutherford_jaxmarl_2024} and CleanRL~\cite{huang_cleanrl_2022}. Table~\ref{tab:comparison} summarizes the comparison; its final column marks that our method, like all listed systems, is validated only in simulation.

\begin{table}[!htb]
\caption{Comparison with Related Systems}
\label{tab:comparison}
\centering
\footnotesize
\begin{tabular}{@{}lcccccc@{}}
\toprule
& \textbf{Co-} & \textbf{Jay-} & \textbf{UQ} & \textbf{GPU} & \textbf{Ped} & \textbf{Real} \\
\textbf{System} & \textbf{train} & \textbf{walk} & \textbf{Metric} & \textbf{Accel} & \textbf{RL} & \textbf{Eval} \\
\midrule
Zhang~\cite{zhang_decision-making_2025} & \texttimes & \checkmark & \texttimes & \texttimes & \texttimes & \texttimes \\
Khuzam~\cite{khuzam_impact_2025} & \texttimes & \checkmark & \texttimes & \texttimes & \texttimes & \texttimes \\
Yu~\cite{yu_multi-agent_2025} & \checkmark & \texttimes & \texttimes & \texttimes & \texttimes & \texttimes \\
Rashid~\cite{rashid_simulation_2024} & \texttimes & \texttimes & \texttimes & \texttimes & \texttimes & \texttimes \\
\textbf{Ours} & \checkmark & \checkmark & \checkmark & \checkmark & \checkmark & \texttimes \\
\bottomrule
\end{tabular}
\end{table}

\begin{figure*}[!t]
\centering
\includegraphics[width=0.85\textwidth]{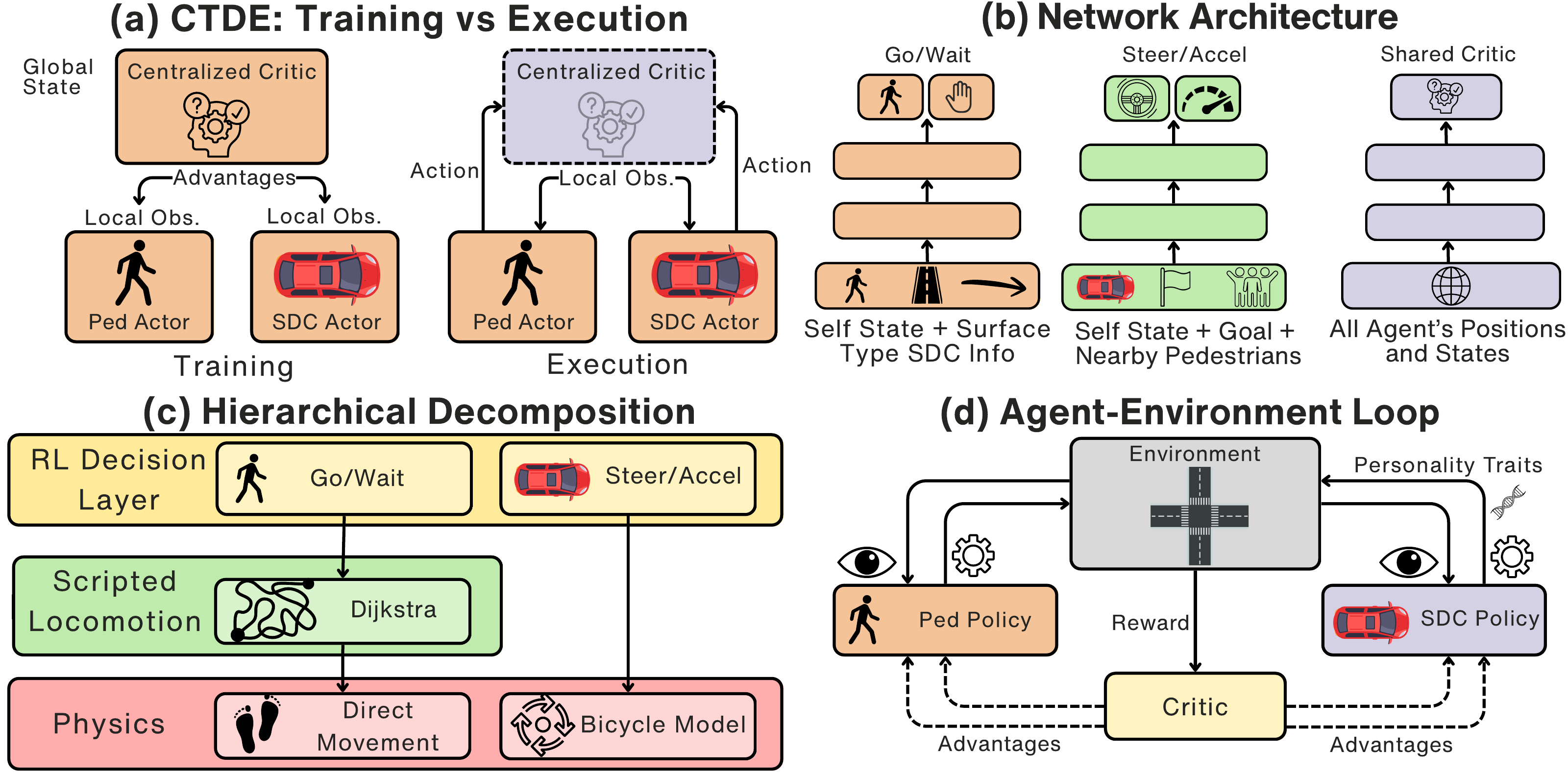}
\caption{System architecture. (a)~CTDE: the centralized critic uses global state during training and is discarded at execution. (b)~Network architecture for pedestrian actor, SDC actor, and shared critic. (c)~Hierarchical decomposition: RL controls go/wait and accel/steer; scripted Dijkstra handles locomotion (SDC bypasses this layer); physics handles motion. (d)~Agent-environment loop with personality traits feeding into the environment.}
\label{fig:sysarch}
\end{figure*}

\section{System Design}

Fig.~\ref{fig:sysarch} overviews the framework: RL policies choose high-level actions (pedestrians go/wait, the SDC accelerates/steers), a scripted layer routes pedestrians along precomputed shortest paths (the SDC bypasses it), and a physics layer advances the bicycle model, all under CTDE training (Fig.~\ref{fig:sysarch}a--d). The environment, training, and inference run in JAX on a single graphics processing unit (GPU) via \texttt{jax.vmap} (512 parallel environments) and \texttt{jax.lax.scan} (rollout collection).

\subsection{Environment}

The environment covers a 120$\times$120\,m urban map with a four-way intersection and a T-junction (3 road segments, 20 sidewalk segments, 6 crosswalks), so a single episode spans both intersection topologies. Simulation runs at $dt{=}0.1$\,s (10\,Hz) for 500 steps (50\,s) per episode.

\textbf{Pedestrians.} The environment contains 12 pedestrian agents, far more than the up to 3 in the recent CARLA-based MARL intersection study of Yu et al.~\cite{yu_multi-agent_2025}, to sustain multiple concurrent pedestrian-SDC interactions per episode. Pedestrians navigate via Dijkstra shortest-path on a 40-node navigation graph (34 sidewalk waypoints plus 6 crosswalk midpoints). The RL policy outputs a binary go/wait decision. When the policy selects ``go,'' a personality-driven roll determines whether the pedestrian crosses via a designated crosswalk or jaywalks across the road:
\begin{equation}
P(\text{jaywalk} \mid \text{go}) = \tau_j \times 0.25
\label{eq:jaywalk}
\end{equation}
where $\tau_j \in [0,1]$ is the jaywalking tendency, sampled uniformly at episode start and \emph{not observable} by the SDC. This trait-based parameterization of pedestrian heterogeneity is in the same spirit as the risk-taking, cautious, and distracted pedestrian types used in prior social-force simulations of AV-pedestrian interaction~\cite{rashid_simulation_2024}. Walking speed ranges from 1.0 to 2.0\,m/s.

\textbf{SDC setting.} The SDC uses a kinematic bicycle model~\cite{polack_kinematic_2017} with wheelbase 2.5\,m, maximum speed 8.33\,m/s (30\,km/h), and maximum steering angle 0.52\,rad. A hard constraint keeps the SDC on the road, applied as a post-physics projection outside the policy and reward: any off-road position is snapped back to within a 0.5\,m road margin and the speed is reduced to 30\%. An episode terminates on collision (SDC-pedestrian centre distance below 1.5\,m), on goal (distance below 3.0\,m), or on timeout.

\subsection{MAPPO Training}

Algorithm~\ref{alg:mappo} summarizes the training loop. We use clipped PPO objectives with GAE and a shared centralized critic, following the common MAPPO design~\cite{yu_surprising_2022}. The number of updates (5{,}000) was chosen empirically by monitoring the learning curves of all three networks until they plateaued on held-out seeds; combined with 512 parallel environments and 256-step rollouts, this corresponds to $6.55 \times 10^{8}$ environment steps in total. The centralized critic observes a 58-dim global state (pedestrian positions, speeds, jaywalking tendencies, and SDC state); the pedestrian and SDC actors use 20-dim and 34-dim local observations (own state, traits or goal, surface or lane information, and nearby agents). All 12 pedestrians share actor parameters, as is standard for homogeneous MARL agents~\cite{yu_surprising_2022}. Each actor and the critic is a two-layer multilayer perceptron (MLP) with ReLU activations: 128 units per layer for the pedestrian actor and 256 for the SDC actor and critic. Table~\ref{tab:setup} lists the full configuration.

\begin{algorithm}[!htb]
\caption{MAPPO Co-Training}
\label{alg:mappo}
\footnotesize
\begin{algorithmic}[1]
\STATE Initialize ped actor $\pi_\theta^p$, SDC actor $\pi_\phi^s$, shared critic $V_\psi$
\FOR{update $= 1$ to $5{,}000$}
\STATE Collect rollouts across 512 parallel envs for 256 steps
\FOR{each env step}
\STATE Ped obs $\gets$ local (20-dim); SDC obs $\gets$ local (34-dim)
\STATE Ped actions $\sim \pi_\theta^p$; SDC action $\sim \pi_\phi^s$
\STATE Step env; record $(o, a, r, d)$
\ENDFOR
\STATE Compute $V(s)$ using critic with global state (58-dim)
\STATE Compute GAE advantages ($\gamma{=}0.995$, $\lambda{=}0.95$)
\STATE Update $\pi_\theta^p$, $\pi_\phi^s$, $V_\psi$ with clipped PPO ($\epsilon{=}0.2$, 4 epochs, 8 minibatches)
\ENDFOR
\end{algorithmic}
\end{algorithm}

\begin{table}[!htb]
\caption{System configuration and hyperparameters. CPU: central processing unit; RAM: random-access memory; NN lib: neural-network library; LR: learning rate; FPS: frames per second; Ent: entropy coefficient.}
\label{tab:setup}
\centering
\footnotesize
\begin{tabular}{@{}ll|ll@{}}
\toprule
\textbf{System} & \textbf{Spec} & \textbf{Param} & \textbf{Value} \\
\midrule
GPU & RTX 5070 Ti 16\,GB & Parallel envs & 512 \\
CPU & Ryzen 7 9700X & Rollout len & 256 \\
RAM & 32\,GB DDR5 & PPO epochs & 4 \\
Framework & JAX 0.6.2 & Minibatches & 8 \\
 & + CUDA 12.6 & Updates & 5{,}000 \\
NN lib & Flax 0.10.7 & LR & $3{\times}10^{-4}$ \\
Optimizer & Optax 0.2.8 & $\gamma$ / $\lambda$ & 0.995 / 0.95 \\
Distributions & Distrax & $\epsilon_{\text{clip}}$ & 0.2 \\
Training FPS & ${\sim}$558k & Ent (ped/SDC) & 0.03 / 0.01 \\
Total steps & 655M & Grad norm & 0.5 \\
\bottomrule
\end{tabular}
\end{table}

The critic trains on a blended reward (50\% mean pedestrian, 50\% SDC) for the cooperative objective. Pedestrians are rewarded for waypoint progress ($+2.0\,\Delta d$) and reaching waypoints ($+5.0$), penalized for collisions ($-25.0$), and given a smart-waiting bonus ($+0.3$) near a fast-approaching SDC. The SDC is rewarded for goal progress and arrival ($+50.0$) and penalized for collisions ($-50.0$), speeding near occupied crosswalks and jaywalkers, off-lane driving, and heading misalignment, with no reward for stopping (preventing indefinite yielding).

\subsection{Uncertainty Quantification}

We distinguish two encounter types from the SDC's perspective: \emph{predictable} (crosswalk crossings, anticipatable from crosswalk proximity and trajectory) and \emph{uncertain} (jaywalking, governed by the latent trait $\tau_j$ invisible to the SDC). We quantify uncertainty via a \textbf{Speed Differential} metric. For each timestep where a pedestrian of type $c \in \{\text{cw}, \text{jw}\}$ is within distance bin $[d_1, d_2]$ of the SDC, we record the SDC speed $v_{\text{sdc}}$:
\begin{equation}
\bar{v}_c(d_1, d_2) = \frac{1}{|T_c|}\sum_{t \in T_c} v_{\text{sdc}}(t), \quad T_c = \{t : d_t^c \in [d_1, d_2]\}
\label{eq:speed_diff}
\end{equation}

The gap $\Delta v = \bar{v}_{\text{jw}} - \bar{v}_{\text{cw}}$ measures how much faster the SDC travels near jaywalkers than crosswalk users; a positive $\Delta v$ indicates an unanticipated encounter. We also report \textbf{collision attribution}, the fraction of collisions involving each pedestrian type.

\section{Experiments}

Policies are trained on randomized episodes with personality traits resampled at the start of each episode. All results use 500-episode evaluations on a fixed seed set identical across methods, so every method is compared on the same scenarios. We report goal rate and collision rate, standard safety and success metrics for RL-based autonomous driving evaluation (CARLA uses similar metrics~\cite{dosovitskiy_carla_2017}). Scenario seeds in Fig.~\ref{fig:intro_scenarios} were selected by fixed programmatic criteria on encounter proximity rather than hand-picked.

\subsection{Baseline Comparison}

\begin{figure}[!htb]
\centering
\includegraphics[width=\columnwidth]{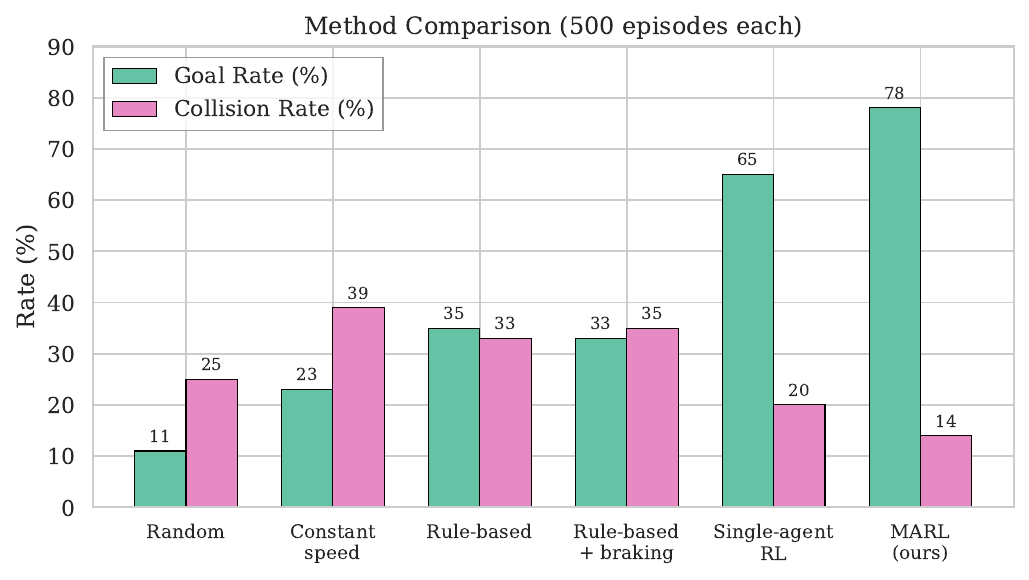}
\caption{Goal and collision rates across methods (500 episodes each). Single-agent RL was trained with scripted pedestrians.}
\label{fig:baselines}
\end{figure}

Fig.~\ref{fig:baselines} compares the MARL SDC against four non-learning baselines and a single-agent RL SDC over 500 episodes. The four non-learning baselines are \emph{random} (uniform acceleration/steering), \emph{constant speed} (fixed throttle, no steering or awareness), \emph{rule-based} (full throttle steering toward the goal), and \emph{rule-based + braking} (rule-based plus braking when a pedestrian is ahead), reaching 11\%/25\%, 23\%/39\%, 35\%/33\%, and 33\%/35\% (goals/collisions); rule-based full throttle is best, and braking does not help because it cannot steer around pedestrians. The co-trained MARL SDC reached 78\%/14\%, and the single-agent RL SDC reached 65\%/20\%.

\subsection{Uncertainty Results}

\begin{figure}[!htb]
\centering
\includegraphics[width=\columnwidth]{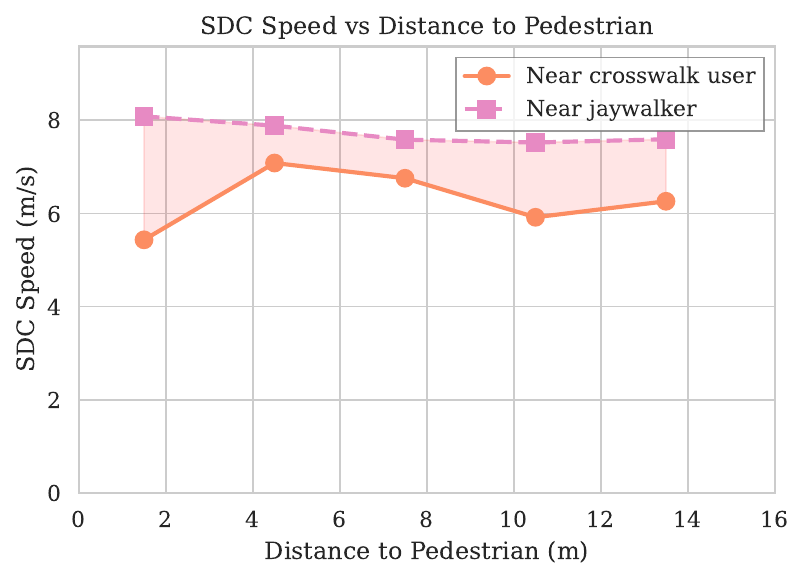}
\caption{SDC speed vs.\ distance to nearest pedestrian, separated by encounter type. The shaded region indicates the speed gap between jaywalker and crosswalk encounters.}
\label{fig:speed_diff}
\end{figure}

Fig.~\ref{fig:speed_diff} shows the speed differential from~\eqref{eq:speed_diff}. At 0--3\,m, the SDC was 2.65\,m/s faster near jaywalkers (8.08 vs.\ 5.43\,m/s), and the gap persists across all distance bins (0.80--1.60\,m/s at 3--12\,m). Near crosswalk users the SDC had already decelerated, while near jaywalkers it maintained a higher speed, consistent with the crossing being unanticipated.

\textbf{Collision attribution.} Jaywalking accounted for 13\% of crossing events but was associated with 62\% of collisions. Two further indicators corroborate the speed differential: the 5th-percentile minimum approach distance was 3.25\,m for jaywalker encounters versus 3.49\,m for crosswalk encounters, and the mean time from a jaywalker entering the road to the SDC initiating braking was 3.38\,s (33.8 steps).

\textbf{Personality mapping.} Jaywalking rate increases monotonically with the trait: Q1 ($\tau_j < 0.25$) $=$ 3.2\%, Q2 $=$ 9.4\%, Q3 $=$ 15.7\%, Q4 ($\tau_j > 0.75$) $=$ 21.6\%, consistent with~\eqref{eq:jaywalk}.

\subsection{MARL vs.\ Single-Agent RL}

Table~\ref{tab:marl} presents a 2$\times$2 comparison. Co-training reduces collisions from 20\% to 14\%. The pedestrian RL policy learns to wait when the SDC is nearby and moving fast, whereas scripted pedestrians never wait. The co-trained SDC also generalizes well against scripted pedestrians (76\% vs.\ 65\% for the single-agent SDC).

\begin{table}[!htb]
\caption{MARL vs.\ Single-Agent RL (500 episodes each)}
\label{tab:marl}
\centering
\footnotesize
\begin{tabular}{@{}lcc@{}}
\toprule
\textbf{SDC $\rightarrow$ Ped Setting} & \textbf{Goal\%} & \textbf{Col\%} \\
\midrule
Single-agent $\rightarrow$ Always go & 65 & 20 \\
Single-agent $\rightarrow$ MARL peds & 65 & 16 \\
MARL $\rightarrow$ Always go & 76 & 16 \\
\textbf{MARL $\rightarrow$ MARL peds} & \textbf{78} & \textbf{14} \\
\bottomrule
\end{tabular}
\end{table}

\subsection{Ablation: Jaywalking Rate}

We scaled the jaywalking probability via the $0.25$ multiplier in Eq.~\eqref{eq:jaywalk}, giving effective jaywalking rates of 0\%, 5\%, 13\% (default), 30\%, and 50\%, with goal/collision rates of 77\%/15\%, 77\%/15\%, 76\%/16\%, 73\%/18\%, and 64\%/28\% respectively. Performance degrades gracefully then sharply: the collision rate rises only from 15\% to 18\% between 0\% and 30\% jaywalking, then jumps to 28\% at 50\%, indicating the SDC tolerates a realistic amount of unpredictable crossings but degrades non-linearly once a large fraction of pedestrians bypass the crosswalks, consistent with jaywalking acting as a tunable source of aleatoric uncertainty.

\section{Discussion and Conclusion}

The speed differential metric gives a trajectory-level measure of uncertainty that complements human-perception-based SDC assessment~\cite{birchler_how_2024}. The realism of the pedestrian model shapes which situations an SDC faces: a scripted-pedestrian testbed underrepresents jaywalking and its surprise ($62\%$ of collisions from $13\%$ of crossings), whereas co-trained, trait-driven pedestrians expose the SDC to more latent-intent situations and yield a metric computable for any policy under test. Co-training is a tool for generating and stress-testing realistic scenarios, not a claim that real pedestrians optimize a joint objective; likewise, the 14\% collision rate is a relative-improvement result under deliberately adversarial jaywalking, not an absolute safety guarantee, and a deployment would add explicit safety layers such as control barrier functions~\cite{ARYAN2026114379}.

Our evaluation is limited to a single synthetic map and a fixed seed set, so generalization to unseen layouts and traffic densities is the primary next step. Future work includes cross-map and varied-density evaluation, comparison against stronger planning- and risk-aware baselines, validation of the speed-differential metric against real-world data, a speed-based collision criterion, per-agent-type critics, human-in-the-loop scenario validation \cite{birchler_how_2024}, and scaling to larger heterogeneous traffic.

\bibliographystyle{IEEEtran}
\bibliography{references}

\end{document}